%% file: main.tex
\documentclass[10pt,twocolumn,letterpaper]{article}

\usepackage{graphicx} 
\usepackage{cuted}
\usepackage{lipsum}
\usepackage{amsfonts}
\usepackage{xcolor}
\usepackage{soul}
\usepackage{comment}
\usepackage{float}
\usepackage{array}
\usepackage{multirow}
\usepackage{authblk}
\usepackage{svg}

\usepackage{iccv}      


\definecolor{iccvblue}{rgb}{0.21,0.49,0.74}
\usepackage[pagebackref,breaklinks,colorlinks,allcolors=iccvblue]{hyperref}

\def\paperID{0015} 
\def\confName{ICCV}
\def\confYear{2025}

\title{Modeling Time-Lapse Trajectories to Characterize Cranberry Growth}

\author{
Ronan John\quad
Anis Chihoub\quad
Ryan Meegan\quad
Gina Sidelli\quad
Jeffery Neyhart\quad
Peter Oudemans\quad
Kristin Dana\quad
}

\affil{Rutgers University - New Brunswick}

\begin{document}
\maketitle
\input{sec/0_abstract}    
\input{sec/1_intro}
\input{sec/2_related_work}

\input{sec/3_dataset}
\input{sec/4_method}
\input{sec/5_experiments}
\input{sec/6_conclusion}
\input{sec/7_acknowledgements}
{
    \small
    \bibliographystyle{ieeenat_fullname}
    \bibliography{main}
}

\end{document}

%% file: sec/0_abstract.tex
\begin{abstract}
Change monitoring is an essential task for cranberry farming as it provides both breeders and growers with the ability to analyze growth, predict yield, and make treatment decisions. However, this task is often done manually, requiring significant time on the part of a cranberry grower or breeder. Deep learning based change monitoring holds promise, despite the caveat of hard-to-interpret high dimensional features and hand-annotations for fine-tuning. To address this gap, we introduce a method for modeling crop growth based on fine-tuning vision transformers (ViTs) using a self-supervised approach that avoids tedious image annotations. We use a two-fold pretext task (time regression and class prediction) to learn a latent space for the time-lapse evolution of plant and fruit appearance. The resulting 2D temporal tracks provide an interpretable time-series model of crop growth that can be used to: 1) predict growth over time and 2) distinguish temporal differences of cranberry varieties. We also provide a novel time-lapse dataset of cranberry fruit featuring eight distinct varieties, observed 52 times over the growing season (span of around four months), annotated with information about fungicide application, yield, and rot. Our approach is general and can be applied to other crops and applications (code and dataset can be found at \url{https://github.com/ronan-39/tlt/}).
\end{abstract}

%% file: sec/1_intro.tex
\section{Introduction}
\label{sec:intro}

\begin{figure}[t!]
\centering
\includegraphics[width=0.4\textwidth]{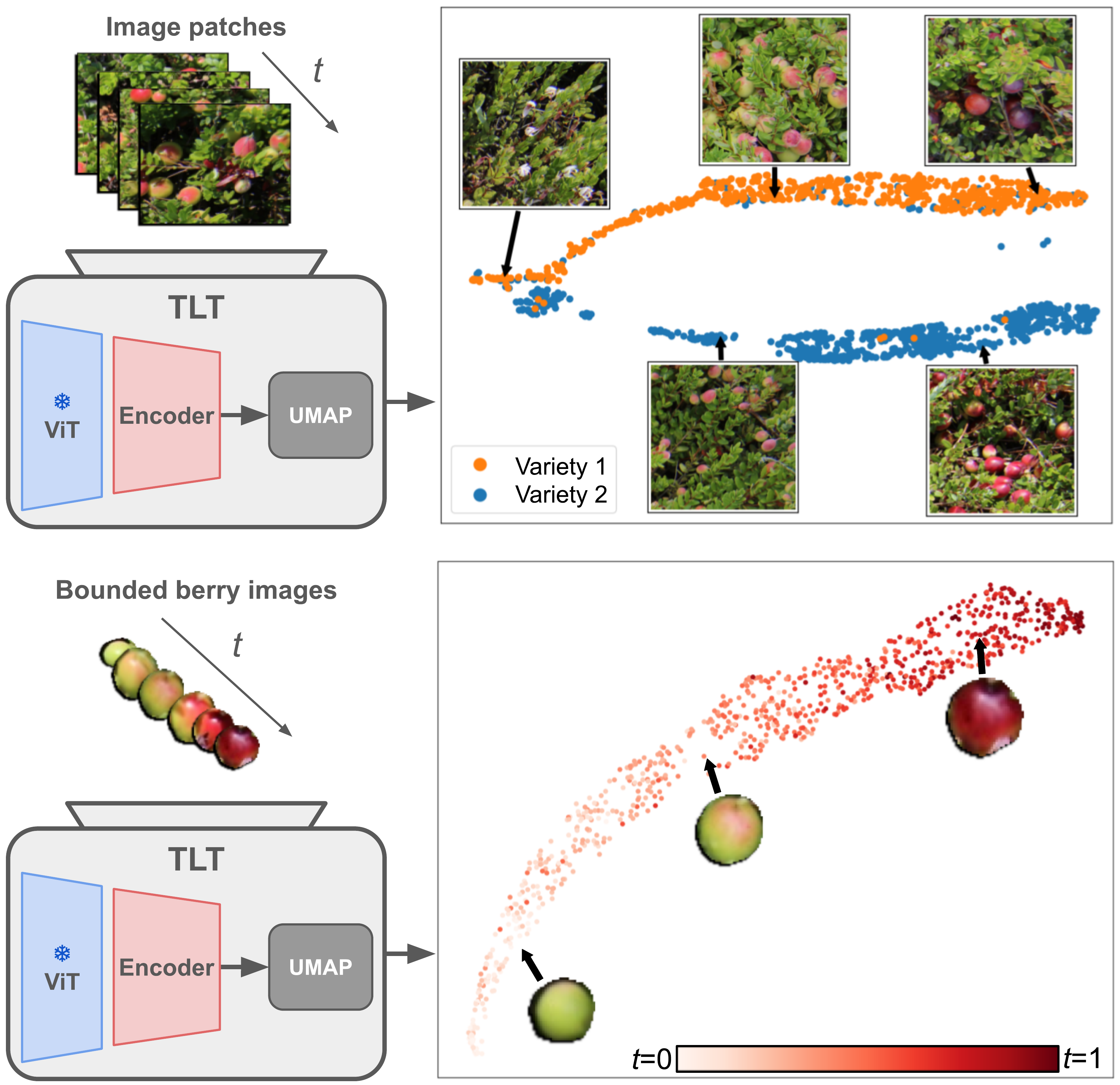}
\caption{In our proposed time-lapse trajectories (TLT) method, image features are projected to an interpretable lower dimensional space that is organized by meaningful differences in crop attributes. This projection is learned by training for several pretext tasks.}
\label{fig:teaser}
\end{figure}

\begin{figure*}[t!]
\centering
\includegraphics[width=0.73\textwidth]{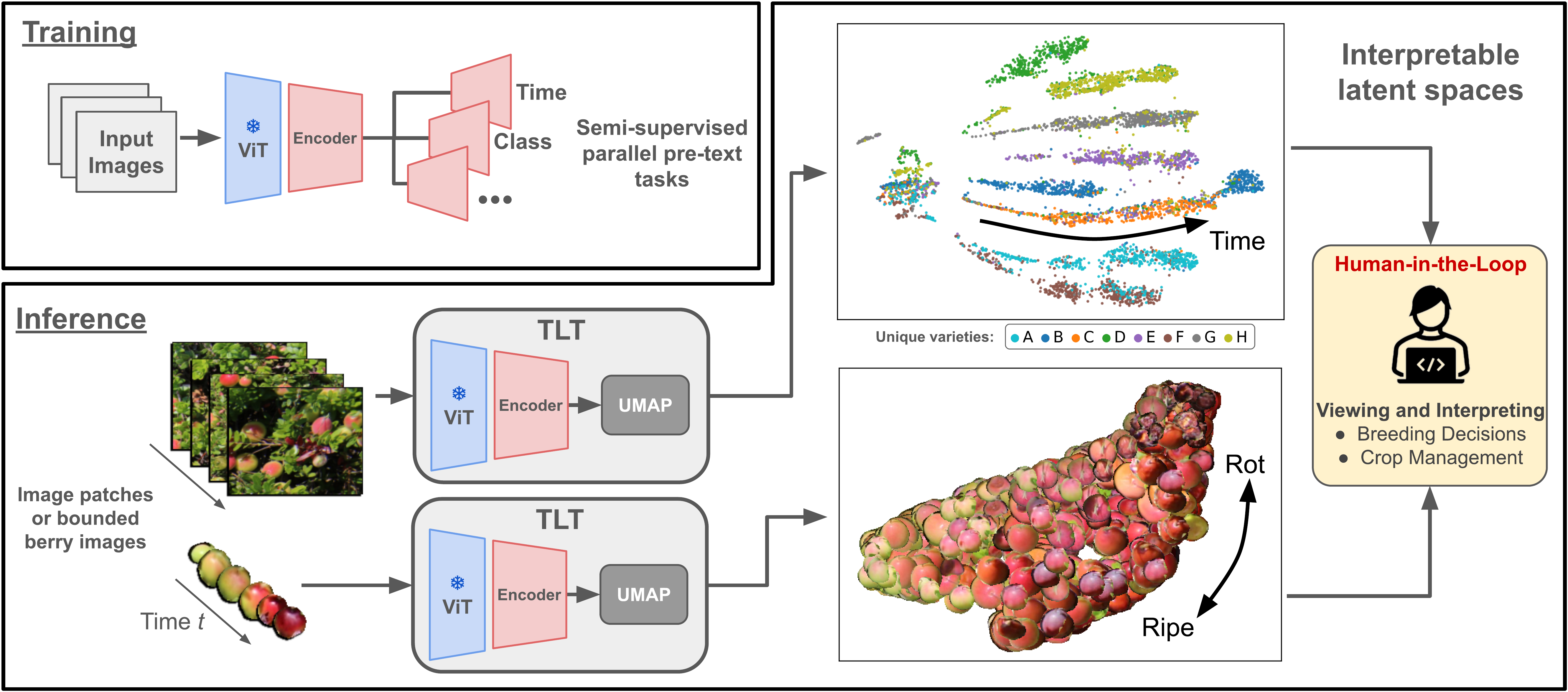}
\caption{Overview of our time-lapse-trajectory method. During training, a frozen pre-trained feature extractor backbone is appended with an encoder, which is jointly trained with several prediction heads for pretext tasks. This encoder is used in conjunction with UMAP to project images into a space that preserves relationships between time, class, etc., based on selected pretext tasks. The temporal tracks for patches are plotted as dots in latent space, while the temporal tracks for berries are plotted with segmented berries shown in the berry-based latent space. Reducing features to 2 dimensions provides interpretability, enabling growers to make informed decisions about breeding and crop management.}
\label{fig:full_method}
\end{figure*}


Quantifying nuanced crop development is critical in agriculture. Growers must efficiently manage their resources to maximize yields, adjusting irrigation in response to temperature fluctuations, timing treatments to prevent the spread of disease, and more. Breeders must monitor trait dynamics—such as ripening rates, growth curves, and stress onset—and associate these phenotypes with genetic background in order to accelerate varietal improvement. Change monitoring—via time-lapse imaging—supports these goals. However, it is labor-intensive, demands specialized expertise, and depends on complex data pipelines. Developing a scalable change-monitoring technique is therefore critical to translate its promise into widespread practice and drive real-world gains in yield and cultivar improvement. 

Scalable change monitoring can be achieved through time-series image acquisition—either via fully autonomous platforms or simple, minimal-cost, user-friendly capture methods—coupled with computer-vision pipelines that forgo labor-intensive expert analysis in favor of objective, quantitative metrics that reliably detect subtle visual cues \cite{meshram2021machine, chiu2020agriculture, zheng2019cropdeep}. For cranberry growers and breeders specifically, change monitoring should focus on mapping plant growth and fruit ripening trajectories, crop responses to treatment regimens, signs of stress, and variety-specific fruit rot and disease susceptibilities \cite{oudemans1998cranberry, polashock2009north, jeffers1991seasonal, averill1997flooding}. Previous works have trained models or deployed foundational models to predict a singular change metric, such as detecting and predicting fruit rot or quantifying ripeness \cite{azim2024, AKIVA2022107444, johnson2025agtech}. However, models that consider only a single metric fail to capture the complexities of change in cranberry plants, substantially constraining the insights that can be drawn. Developing models that jointly integrate multiple change metrics provides a far more complete picture for growers and breeders.

To unify multiple change metrics, we develop a time-lapse trajectory (TLT) framework that leverages vision-transformer foundation models for feature extraction and produces high-dimensional descriptors of diverse visual attributes. We fine-tune the transformer with a specialized encoding layer, trained on pretext tasks, to adapt the features for monitoring changes in cranberry crops. A key insight in our approach is a two-stage projection from the high-dimensional foundation model features to a mid-range latent space that is tuned with quantitative pretext tasks (see Figures~\ref{fig:teaser} and ~\ref{fig:full_method}). A subsequent projection to a two-dimensional latent space transforms the spatio-temporal feature space into comprehensible temporal tracks in the learned 2D latent space. In this space, end users can readily analyze the tracks, comparing predictions to current observations among varieties. The input time series is a time-lapse image sequence of the same spatial region over the growing season obtained from a fiducial marked region (see Figure~\ref{fig:raw_patch_with_pvc}), processed as patches or as segmented berries depending on the desired scale. The resulting latent space trajectories are low-dimensional and interpretable, enabling actionable insights for growers and breeders. In summary, our contributions are: 

\begin{enumerate}[noitemsep]
    \item {\bf TLT}: A framework that models crop growth by learning latent representations from a time series of crop images captured at fixed spatial locations. 
    \item  {\bf TLT prediction module} that forecasts crop development through time-series observations, conditioned on variety-specific cranberry dynamics. 
    \item  {\bf TLT analysis module} for breeders that provides an expected temporal track for a set of cultivars to reveal any positive or negative deviations from desired phenotypes. 
    \item {\bf TLC: Time-lapse Cranberry Dataset} A publicly available dataset imaging 8 cranberry varieties over the course of one growing season (span of around 4 months), annotated with information about fungicide treatment, fruit rot prevalence, and yields.

\end{enumerate}

%% file: sec/2_related_work.tex
\section{Related Work}
\label{sec:related_work}

\paragraph{\bf Growth Modeling and Assessment}
Driven by recent advancements in data collection and deep learning, modeling plant growth is an active area of research. For example, a recent framework \cite{li2025deepchangemonitoringhyperbolic} leverages hyperbolic networks and an annotated tree-cover dataset to learn change from overhead imagery—achieving SoTA results but relying on abundant remote-sensing data. Other methods leverage generative based methods, such as diffusion \cite{ho2020denoising}, GANs \cite{goodfellow2020generative}, and variational autoencoders \cite{pinheiro2021variational}, to model plant growth. For example, GAN networks \cite{transgrow} have been used to create original images depicting seasonal plant growth. In follow up work \cite{Hohldeep2024}, a pre-trained autoencoder is used instead of an end-to-end network \cite{arjovsky2017wassersteingan}. However, these generative models demand hours of training on powerful, memory-intensive hardware and often leave visual artifacts, limiting their practical application.

Cranberry specific growth modeling and management computer vision techniques have explored rot prediction and detection and ripening analysis independently. For example, in \cite{azim2024} the authors used a CNN to distinguish healthy berries from rotten berries based on visual features. Other works have used drone-based imagery and stratified random sampling (images from the same bog but not the same location) to predict berry-rot risk \cite{akiva2021ai, AKIVA2022107444}.
In \cite{johnson2025agtech}, the drone-based imagery dataset is used to quantify cranberry ripening rates and compare cultivars. While these methods provide pioneering steps in cranberry assessment, they lack time-lapse imagery and a holistic analysis beyond ripening and berry counts.

\paragraph{\bf Vision Foundation Models in Agriculture}
Transformer-based foundation models have become an important part of vision-based pipelines. Seminal work \cite{dosovitskiy2021imageworth16x16words} introduced vision transformers (ViTs) that linearly encode patches of an image, and adapted language-based transformers \cite{vaswani2017attention} to vision-based tasks. Modern foundation models are trained on large datasets acquired from publicly available databases, making their features more expressive compared to traditional CNN feature extractors. Since their introduction, there have been several adaptations to ViTs, including DINO models \cite{caron2021emergingpropertiesselfsupervisedvision, oquab2024dinov2learningrobustvisual} and vision-language models \cite{radford2021learningtransferablevisualmodels, zhai2023sigmoidlosslanguageimage, tschannen2025siglip2multilingualvisionlanguage}. Vision foundation models have become a valuable tool in prevision agriculture by providing robust, pre-trained representations that can be effectively adapted for specialized agricultural tasks. These models enable scientists to leverage visual features for applications including disease identification \cite{jiang2025plantcafo, barman2024vit, borhani2022deep}, growth stage classification \cite{jeon2025hybrid, darlan2025smartberry, ulukaya2025robust, johnson2025agtech}, and yield predictions \cite{ge2025winter, guo2024novel, lin2023mmst} even when working with limited domain-specific datasets. Recent work \cite{Chenadapting2023} uses ViT for Cassava leaf segmentation, counting, and disease classification. Another example, \cite{10350922} utilizes the Swin transformer \cite{liu2022swintransformerv2scaling} and VOLO \cite{yuan2021volovisionoutlookervisual} to predict yield of wheat varieties.
\vspace{-0.1in}

\paragraph{\bf Explainable AI in Agriculture}
Interpretability and explainability are important aspects of modern AI systems \cite{xu2019explainable, dwivedi2023explainable}. In applied agriculture, AI adoption may stall until growers and breeders, many of whom may be unfamiliar with machine learning and therefore naturally skeptical, can see exactly how visual evidence translates into objective crop assessments. To improve the explainability of computer vision models, numerous methods have emerged over the last ten years. For visual explainability, a seminal work in this field is Grad-CAM \cite{selvaraju2016grad}, which aims to identify the parts of an image that are the most descriptive for image classification in a CNN. Grad-CAM generates visual explanations for CNN-based models by computing the gradient of the target class score with respect to the feature maps of a convolutional layer. These gradients are used to weight the feature maps and produce a coarse localization map of important regions in the input image. Grad-CAM methods have been adapted to work with the attention maps of ViTs as well. A useful approach to interpretability is 2D visualization of latent spaces to support human-in-the-loop paradigms. T-SNE \cite{van2008visualizing} and UMAP \cite{mcinnes2018umap, mcinnes2020umapuniformmanifoldapproximation} provide excellent 2D manifolds, with UMAP having the advantage of speed and global structure preservation.
\vspace{-0.1in}

%% file: sec/3_dataset.tex
\section{Time-Lapse Dataset}
\label{sec:dataset}

\begin{figure}
    \centering
    \includegraphics[width=1.0\linewidth]{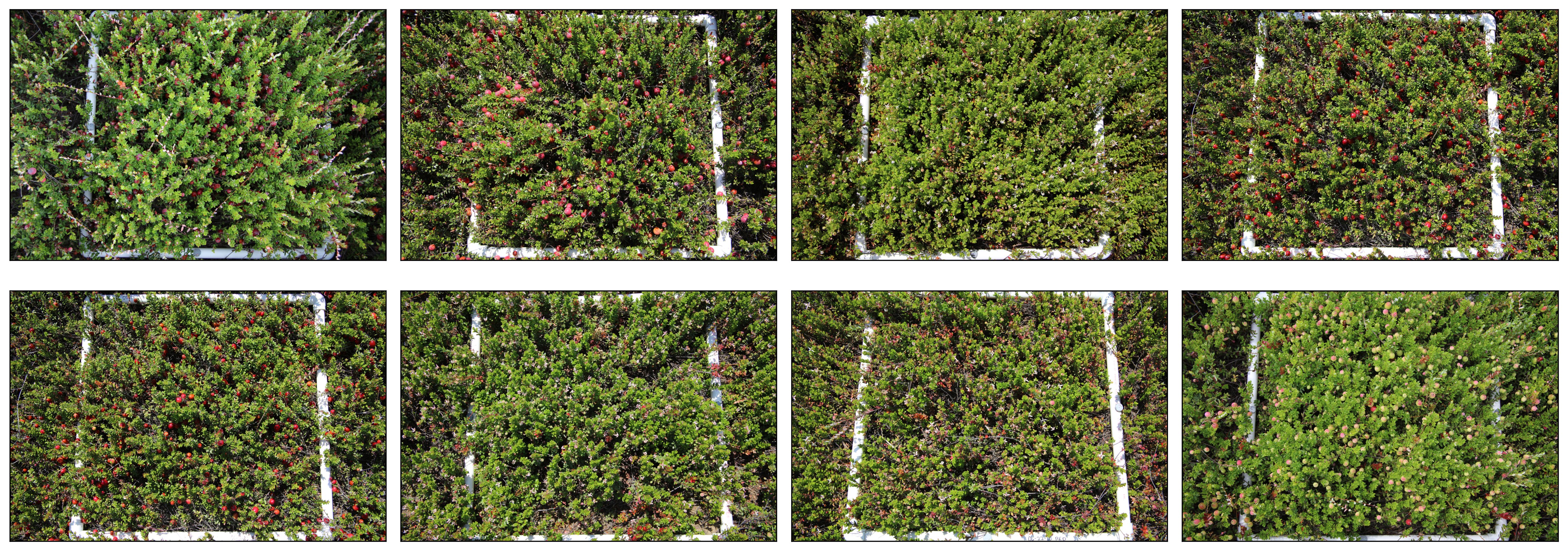}
    \caption{Example region from cranberry bog delineated with a PVC frame fiducial marker for repeatable imaging (filtered out in pre-processing).}
    \label{fig:raw_patch_with_pvc}
\end{figure}

\begin{figure*}[t]
    \centering
    \includegraphics[width=0.65\linewidth]{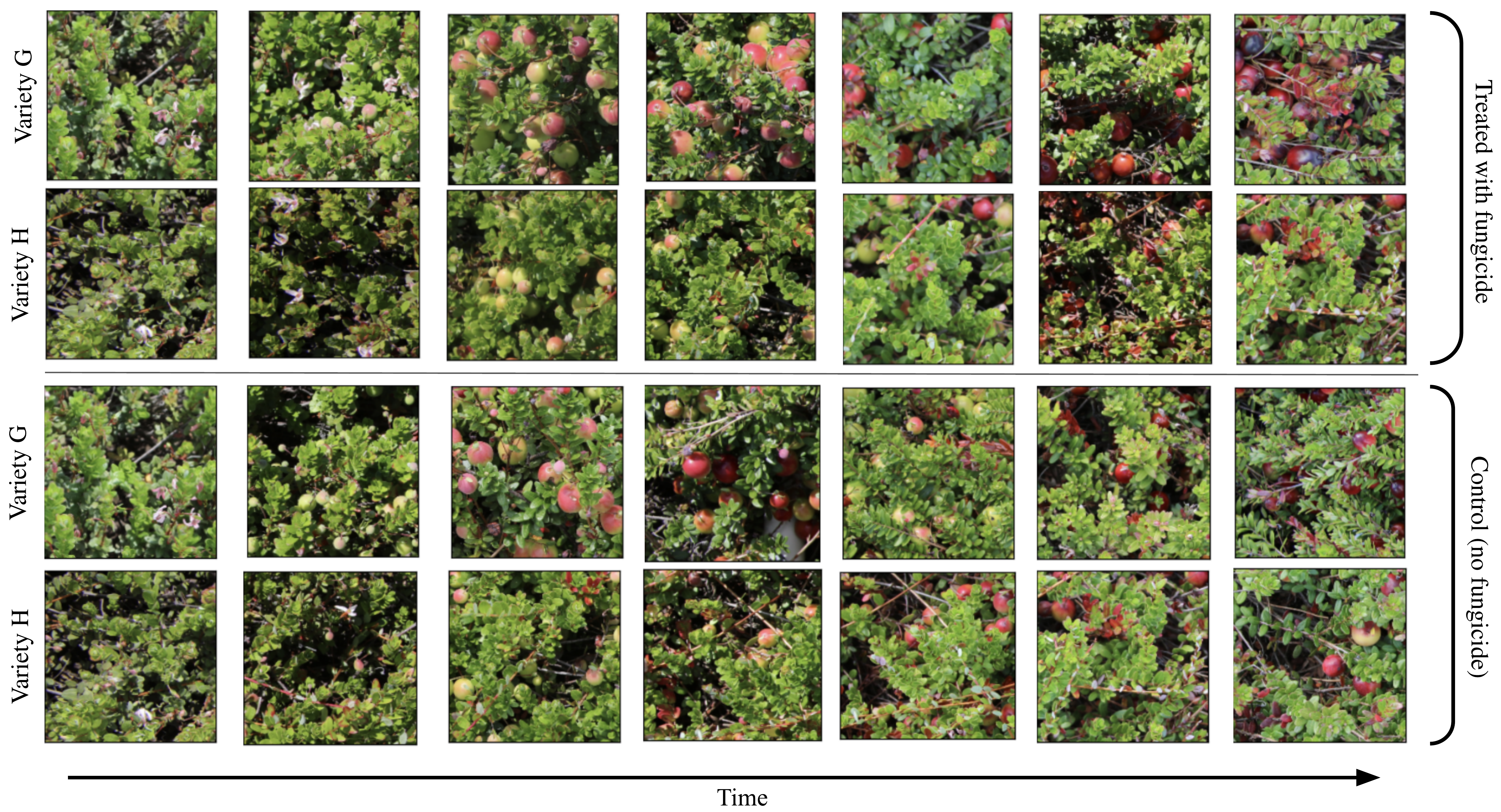}
    \caption{Example images from the TLC (time-lapse cranberry) dataset comprised of 16 delineated regions imaged on 52 dates spanning 108 days with 8 cranberry varieties, each with and without fungicide.}
    \label{fig:Dataset_Examples}
\end{figure*}

\begin{figure}[]
    \centering
    \includegraphics[width=1.0\linewidth]{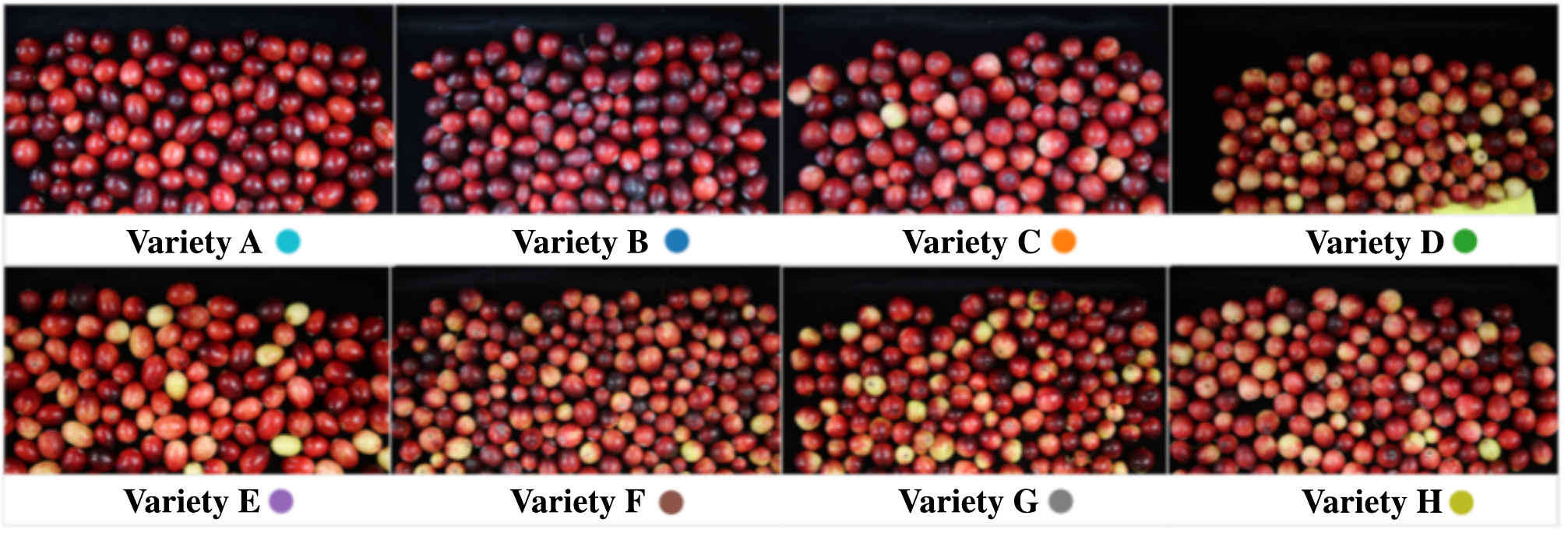}
    \caption{Examples of harvested berries from the 8 different varieties in our dataset. Each variety is color coded in this paper.}
    \label{fig:genotype examples}
\end{figure}

Although quantifying appearance in individual images provides useful insights, modeling those appearance changes over a time series delivers far greater impact for growers and breeders. To support such change monitoring work, we have collected a time-lapse image dataset of regions in cranberry bogs. Our dataset, {\it Time-lapse Cranberry Dataset} (TLC), consists of imagery of one of sixteen regions of cranberry shrubs, each marked by a labeled semi-permanent PVC frame, from approximately the same viewpoint. These sixteen regions were imaged using a hand-held DSLR camera, at a resolution of 8688$\times $5792, for 52 sessions over the span of 108 days (early June to mid September). To mitigate lighting variations between imaging sessions, a Macbeth Color Checker was photographed at the beginning of each session for photometric calibration. Each region corresponds to one of eight distinct cranberry breeds and one of two treatments—fungicide and no fungicide—which resulted in varying levels of fruit rot. Example images from the eight varieties are shown in Figure \ref{fig:genotype examples} and will be referred to by letters given in the figure. In total, we provide 52 images per cranberry breed (8) and treatment (2), resulting in a total of 832 images. Raw images and JPEG files are provided for all of these images. Table \ref{table:dataset_split} presents basic statistics on our dataset.

While existing datasets such as the CRAID dataset introduced in \cite{Akiva_2020_CVPR_Workshops} and the Wild Berry image dataset in \cite{Riz_2025} provide a significant amount of time-series data, these datasets are not designed for change monitoring. For example, CRAID uses drone sampling, imaging the same bog over time, but not the exact same spatial region. A similar set up is seen in the Wild Berry image dataset, where selected plants are imaged over time, but not the same region. Our dataset instead follows a time-lapse approach, where the same region, delineated by a labeled PVC square (see figure \ref{fig:raw_patch_with_pvc}), is imaged over time. This method enables tracking of crop patch and individual berries over time, allowing for detailed observation of ripening rates, treatment efficacy, effects of temperature swings, or early signs of fungal disease. See figure \ref{fig:Dataset_Examples} for some examples of change in image patches over time.

We analyzed the data at both patch and berry scales. In the patch-based approach, we quantify changes within the imaged region. This region encompasses cranberries, leaves, twigs, and other distinct visual elements, building the comprehensive appearance of the plant that manifests genetic traits. This patch-based analysis is particularly effective when comparing varieties and is specifically tailored to the needs of breeders. On the other hand, the berry-based approach involves segmenting and tracking individual berries across time, focusing purely on changes within unique berries (e.g. ripening rates, size, and color). Berry-based analysis will be particularly effective for observing berry growth: how the berries are responding to their environment and or treatments. Therefore, berry-based analysis is more useful when trying to maximize yield, and is specifically tailored to the needs of growers. 

For the patch-based approach, each region image is divided into 224$\times$224 pixel patches (after pre-processing to remove the PVC pipe fiducial marker). For the berry-based approach, we adopted a similar method in \cite{johnson2025agtech}. We align each subset of date-ordered time-lapse images by first extracting XFEAT \cite{potje2024xfeatacceleratedfeatureslightweight} descriptors between adjacent time-series pairs, using those descriptors to generate LighterGlue \cite{lindenberger2023lightgluelocalfeaturematching} keypoint correspondences, estimating homographies from the matched keypoints, and then warping each image’s perspective to match its predecessor. We opted to manually segment 44 time-series berries—each tracked over sequential time points—using point-click inputs to the Segment Anything 2 image predictor class \cite{ravi2024segment} across three cranberry varieties (C, F, and G) and two treatments, in order to avoid inaccuracies that automated methods (e.g. SAM 2 Video tracking) occasionally introduced. These berries were selected because they remained fully visible throughout the time-series imagery, starting green (unripe), and ending either crimson (ripe) or showing rot (e.g. shriveling, discoloration). Rot status was assessed per berry image on a binary basis: rotten or not rotten. Pre‑processing yielded 34 ripe and 10 rotten berry time series (1,456 images: 135 rotten, 1,321 not). We analyzed only varieties C, F, and G because other varieties developed dense canopy growth that hindered consistent berry tracking. These three also capture the phenotypic diversity of all eight varieties, with D–H and A–C forming two visually similar groups.


\begin{table}[h!]
\centering
\scalebox{0.85}{
\begin{tabular} {c c c c}
\multicolumn{4}{c}{\textbf{Time-lapse Cranberry Dataset}} \\
\hline
Span & Imaging dates & Varieties & \# Fungicide treatments \\
\hline
108 days & 52 & 8 & 2 \\
\hline
\end{tabular}
}
\caption{Statistics of the TLC dataset. Each image is annotated with the time it was taken, the variety of cranberry it belongs to, and whether or not it received fungicide treatment. In addition, yield and rot statistics were sampled 9 times throughout the season by partially harvesting identically conditioned nearby regions.}
\label{table:dataset_split}
\end{table}

%% file: sec/4_method.tex
\section{Method}
\label{sec:method}

Our proposed Time-Lapse Tracking (TLT) method learns a latent space where meaningful features of crops can be visualized and modeled for use in predicting crop qualities and statistics. The TLT module consists of a pre-trained feature extractor backbone, followed by dimensionality reduction performed by a trained encoder. Dimensionality reduction is guided by pretext tasks, where incorporating relevant crop statistics during training enables the model to extract key visual features associated with crop changes over time. A second untrained dimensionality reduction is performed using UMAP \cite{mcinnes2020umapuniformmanifoldapproximation} to bring features down to $D$ dimensions (to maintain interpretability, we use $D=2$ in this paper). UMAP preserves the spatial relationships observed in the learned latent space while enabling explainability. Our entire architecture is outlined in \cref{fig:full_method}. 

To train the TLT module, the encoder is tasked with performing multiple pretext tasks. We feed image patches or bounded berry images into the feature extractor backbone. The normalized classifier token $f \in \mathbb R^n$ is fed forward into the encoder, which is implemented as a fully connected network. The encoder is composed of two multi-layer perceptrons (MLPs) with a ReLU activations. The encoder layers reduce feature dimensionality to $n/2$, then $n/4$ sequentially. The output of this encoder, $z = e_\phi(f)$, forms our latent space.

For fine-tuning, the output of the encoder is used as the input for multiple pretext tasks. For each pretext task, we append fully connected prediction heads to our model, which are optimized jointly during training. We consider prediction heads for time (relative to the growing season), class (plant variety), and whether or not a given plant was treated with fungicide. For bounded berry images specifically, we also consider the task of predicting if a berry is rotten. We select pretext tasks that are designed to disentangle environmental effects from the crop's response to them. For example, some parts in an image patch may be cast in shadow or have other small deviations, so we seek to learn lighting invariance and ignore nuisance change within the patch. The time prediction task enables learning this invariance by aligning latent vectors encoded from images on the same day that differ only by superficial lighting changes. Additionally, time prediction and class prediction tasks are aimed at learning the key visual differences that distinguish the crop at different stages of growth.

Class prediction tasks, such as fungicide treatment prediction and predicting if a berry is rotten, use a binary cross entropy (BCE) loss (\ref{eq:bce}). Classes are encoded as one-hot vectors. Tasks to predict a continuous value, such as time, use a mean squared error (MSE) loss (\ref{eq:mse}). Continuous values are normalized to a range of 0-1. For a prediction $y$ and label $\hat{y}$, these loss functions are defined as follows:

\vspace{-0.2cm}

\begin{equation}
    \text{MSE} = \frac{1}{N}\sum_{i=1}^{N}(y_{i}-\hat{y}_{i})^2
    \label{eq:mse}
\end{equation}

\vspace{-0.5cm}

\begin{equation}
    \text{BCE} = -\frac{1}{N}\sum_{i=1}^{N}\Bigl[\,y_i \,\log \hat{y}_i \;+\; (1 - y_i)\,\log\bigl(1 - \hat{y}_i\bigr)\Bigr]
    \label{eq:bce}
\end{equation}

The final loss function sums the individual losses for each prediction head:
\begin{equation}
    \mathcal{L_\text{total}} = \mathcal{L_{\text{time}}} + \mathcal{L_{\text{variety}}} + \mathcal{L_{\text{fungicide}}},
\end{equation}
where $\mathcal{L_\text{time}}$ is an MSE loss and $\mathcal{L_\text{variety}}, \mathcal{L_\text{fungicide}}$ are BCE losses. All models are trained with the Adam optimizer \cite{kingma2014adam} with a learning rate of 0.005 for 8 epochs using the PyTorch \cite{paszke2019pytorchimperativestylehighperformance} framework.

After obtaining the latent vectors from the trained encoder, UMAP is used to project the latent vector down to two dimensions. In this reduced space, we observe that features from specific varieties follow predictable trajectories over time. We seek to model these trajectories such that we can predict the future state of a variety. We start with the set of points projected to the latent space, $X = \{x_1, x_2, ... x_T\} \in \mathbb R^{D\times T}$. We then calculate the relative position for each point, which we refer to as velocity:
\begin{equation}
    V = \{x_{t+\epsilon} - x_t\ |\ t \in (0, T-\epsilon)\}
\end{equation}
Where $\epsilon$ is a parameter to denoise the velocities, and $T$ is the maximum length of a sequence. We fit a Bayesian Gaussian mixture model to the distribution of the stacked position and velocity vectors: $P([X V])$. Fitting this distribution to the training data effectively obtains time-invariant representations of training trajectories. During inference, a starting point, $x_0$, can be chosen and the distribution can be conditioned to obtain $P(V | X=x_0)$. From here, we repeatedly integrate velocity and update position to sample a likely trajectory.

%% file: sec/5_experiments.tex
\section{Experiments} \label{sec:experiments}

\subsection{Crop Metric Prediction} \label{sec:crop_metric_prediction}
In this section, we evaluate the performance of our pretext tasks, which aim to predict useful metrics in crop age, crop variety, whether or not fungicide has been applied, and if a berry is rotten (for bounded berry images only). All plant varieties are considered for image patches, and three plant varieties are considered (C, F, G) in the bounded berry images. Image patches and bounded berry images are both split into a 70/30 test train split. To avoid memorization while learning, the patch-wise split is constructed such that patches from the left sides of images will only ever appear in the training set, and vise versa for the test set. Each feature extractor backbone is evaluated on three pretext tasks (time, class, and fungicide), with bounded berry images additionally being evaluated on the rot prediction task. Performance is evaluated independently for each prediction head. Time prediction is evaluated with mean absolute error (MAE) in days, and percent agreement (PA) for class, rot, and fungicide treatment prediction. These metrics are defined as follows:

\begin{equation}
    \text{MAE} = \frac{1}{N}\sum_{i=1}^{N}(y_{i}-\hat{y}_{i}),
    PA = \dfrac{100\%}{N}\sum\limits_{i=1}^N[\hat{y_i}=y_i]
    \label{eq:mae}
\end{equation}

Image patch results are considered first. We present the performance baselines with different configurations of prediction heads with each feature extractor backbone in tables \labelcref{table:time_only,table:time_geno_only,table:time_class_fungicide}. All models perform adequately in the time prediction task, seen in \cref{table:time_only}.
 
\begin{table}[h!]
\centering
\begin{tabular}{c c} 
 \hline
  & Time MAE(d) $\downarrow$\\ [0.5ex] 
 \hline
 ViT & $3.51 \pm 3.17$  \\ 
 DINOv2 & $\mathbf{3.39 \pm 3.55}$  \\
 Swin & $5.166 \pm 5.01$ \\
 SigLIP & $4.41 \pm 4.58$ \\  [1ex] 
 \hline
\end{tabular}
\caption{Image patch baseline results for time prediction task with different feature extractor backbones. Best results in bold. MAE is reported in days. The total time span of the dataset is 108 days.}
\label{table:time_only}
\end{table}

\begin{figure}[h!]
\centering
\includegraphics[width=0.9\linewidth]{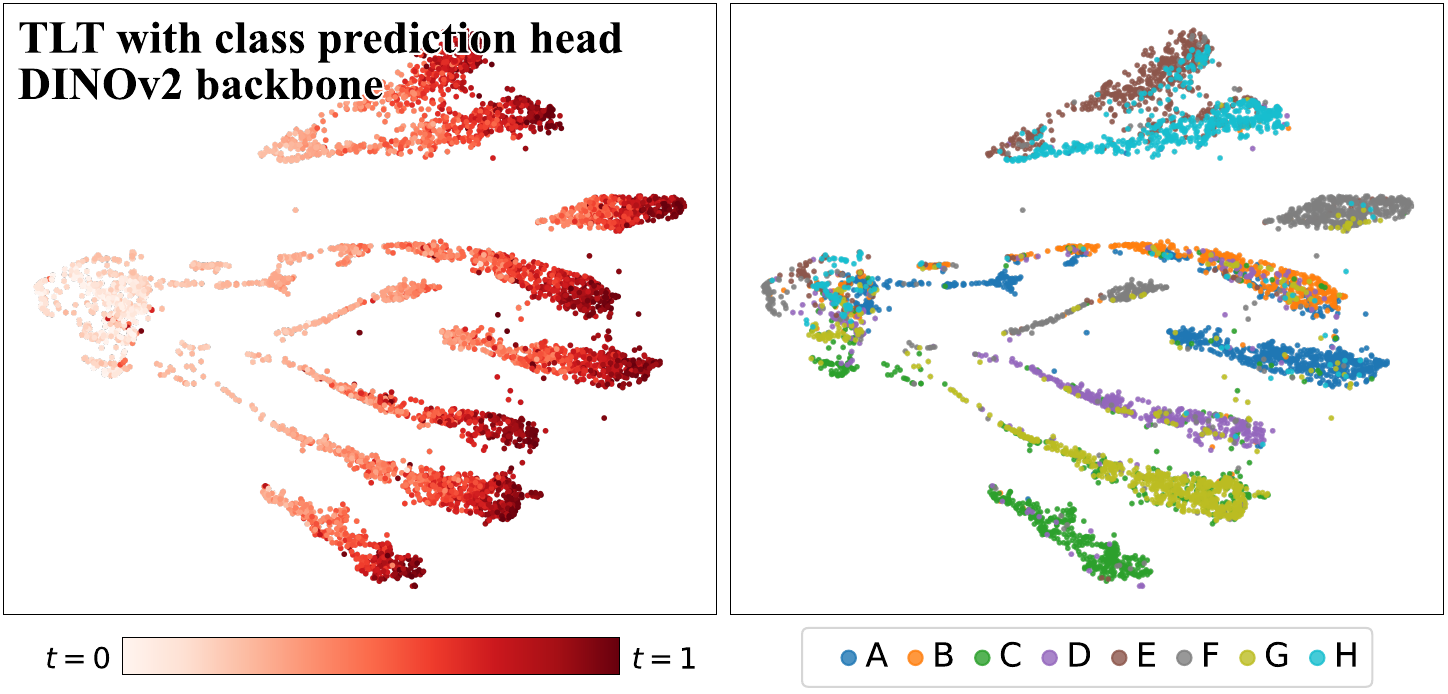}
\caption{Image patch TLT module with time and class heads demonstrates separation by both in latent space. Colored by ground truth time (L) and class (R).}
\label{fig:umap_full_split}
\end{figure}

\begin{figure}[h!]
\centering
\includegraphics[width=0.9\linewidth]{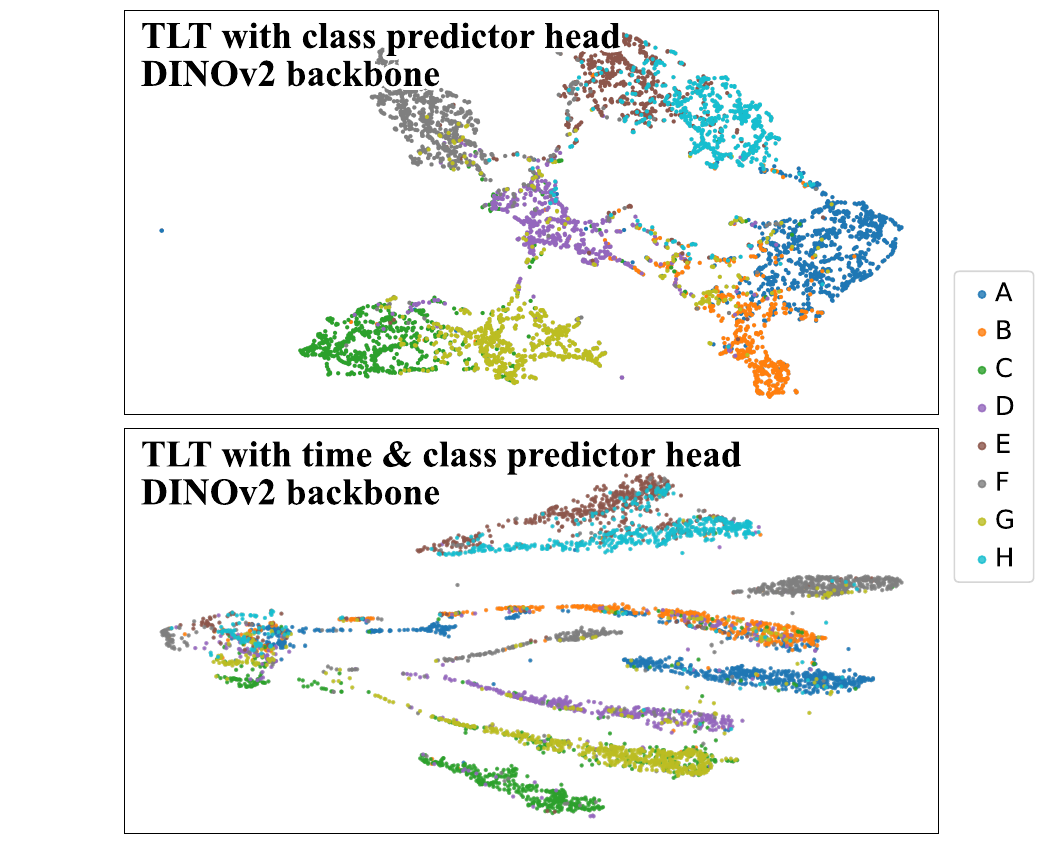}
\caption{Image patch TLT projections of models trained with a class predictor head \textbf{(top)}, vs. a model trained with a time and class predictor head \textbf{(bottom)}. Time prediction as a pretext task organizes the latent space by time, resulting in cleaner trajectories. Colorized by ground truth class.} 
\label{fig:umap_compare_heads}
\end{figure}

When training to predict time and class, time performance is relatively unhampered, as shown in \cref{table:time_geno_only}. DINOv2 is able to correctly predict the correct class 79.4\% of the time. ViT and SigLIP trail in the low 60\%s, and Swin struggles to predict class at 20\%, which is marginally better performance than random guessing (12.5\%). When training with all three prediction heads, the performance of any given task degrades compared to the performance when the model is fine tuned on a single task. The relatively high performance of the time prediction class despite this intuitively suggests that the biggest visible differences are observed across time. The class prediction decreases the most, with the top performer DINOv2 reaching 51.4\%. Fungicide treatment prediction presents strong performance for all backbones. In general, DINOv2 emerges as a particularly strong feature backbone for this set of tasks.  Visualizations of some of these learned latent spaces are shown in \ref{fig:umap_full_split} and \ref{fig:umap_compare_heads}. Similar results were also observed in the bounded berry images. The addition of the rot prediction head had no noticeable impact on the performance of the other heads and the bounded berry image TLT continued to perform strongly regardless of the number of heads that were used (see \cref{table:berry_all_head}). 

\begin{table}[h!]
\centering
\begin{tabular}{c c c} 
 \hline
  & Time & Class \\ [0.5ex] 
  & MAE(d) $\downarrow$ & PA[\%] $\uparrow$ \\ [0.5ex] 
 \hline
 ViT & $\mathbf{3.49}$ & $63.1\%$ \\ 
 DINOv2 & $3.56$ & $\mathbf{79.4\%}$ \\
 Swin & $5.32$ & $24.7\%$ \\
 SigLIP & $4.29$ & $60.1\%$ \\  [1ex] 
 \hline
\end{tabular}
\caption{Image patch results for joint task of time and class prediction tasks with different feature extractor backbones. Best results in bold.}
\label{table:time_geno_only}
\end{table}

\begin{table}[h!]
\centering
\begin{tabular}{c c c c} 
 \hline
  & Time & Class & Fungicide \\ [0.5ex] 
  & MAE(d) $\downarrow$ & PA[\%] $\uparrow$ & PA[\%] $\uparrow$ \\ [0.5ex] 
 \hline
 ViT & $\mathbf{5.65}$ & 38.9\% & 77.5\% \\ 
 DINOv2 & 5.76 & $\mathbf{51.4\%}$ & $\mathbf{84.6\%}$ \\
 Swin & 5.74 & 30.5\% & 80.5\% \\
 SigLIP & 5.70 & 32.6\% & 82.2\% \\  [1ex] 
 \hline
\end{tabular}
\caption{Image patch results for joint task of time, class, fungicide prediction tasks with different feature extractor backbones. Best results in bold. Addition of the fungicide prediction head hampers performance in class prediction.}
\label{table:time_class_fungicide}
\end{table}

\begin{table}[h!]
\centering
\begin{tabular}{c c c c c}
 \hline
  & Time & Class & Fungicide & Rot\\ [0.5ex] 
  & MAE(d) $\downarrow$ & PA[\%] $\uparrow$ & PA[\%] $\uparrow$ & PA[\%] $\uparrow$ \\ [0.5ex] 
 \hline
 ViT & 6.93 & 45.9\% & 62.7\% & 93.5\%\\
 DINOv2 & $\mathbf{5.89}$ & $\mathbf{54.8\%}$ & $\mathbf{64.4\%}$ & $\mathbf{94.8\%}$ \\
 Swin & 7.81 & 46.1\% & 63.7\% & 93.5\% \\
 SigLIP & 6.32 & 45.2\% & 58.3\% & 94.7\% \\  [1ex] 
 \hline
\end{tabular}
\caption{Bounded berry image results for joint task of time, class, fungicide, and rot prediction tasks with different feature extractor backbones. Best results in bold. The rot prediction retained its performance as more heads were added and did not degrade the performance of other heads.}
\label{table:berry_all_head}
\end{table}


\subsection{Latent Space Trajectory Modeling}
To observe the learned latent space of the encoder, we start by taking the patch-based training data used to train a given encoder, and project that training data to the learned latent space. We project these latent features to $2$ dimensions, storing the transformation that was learned with UMAP. We observe that the pretext tasks yield latent spaces where particular varieties follow predictable paths over time in UMAP space. We then model and predict these trajectories. During inference, the test set is projected down with the trained encoder and the UMAP transformation learned from the training set. This projected test set is used to evaluate the qualitative performance and generalization of the learned trajectory.

\begin{figure}[h!]
\centering
\includegraphics[width=1.0\linewidth]{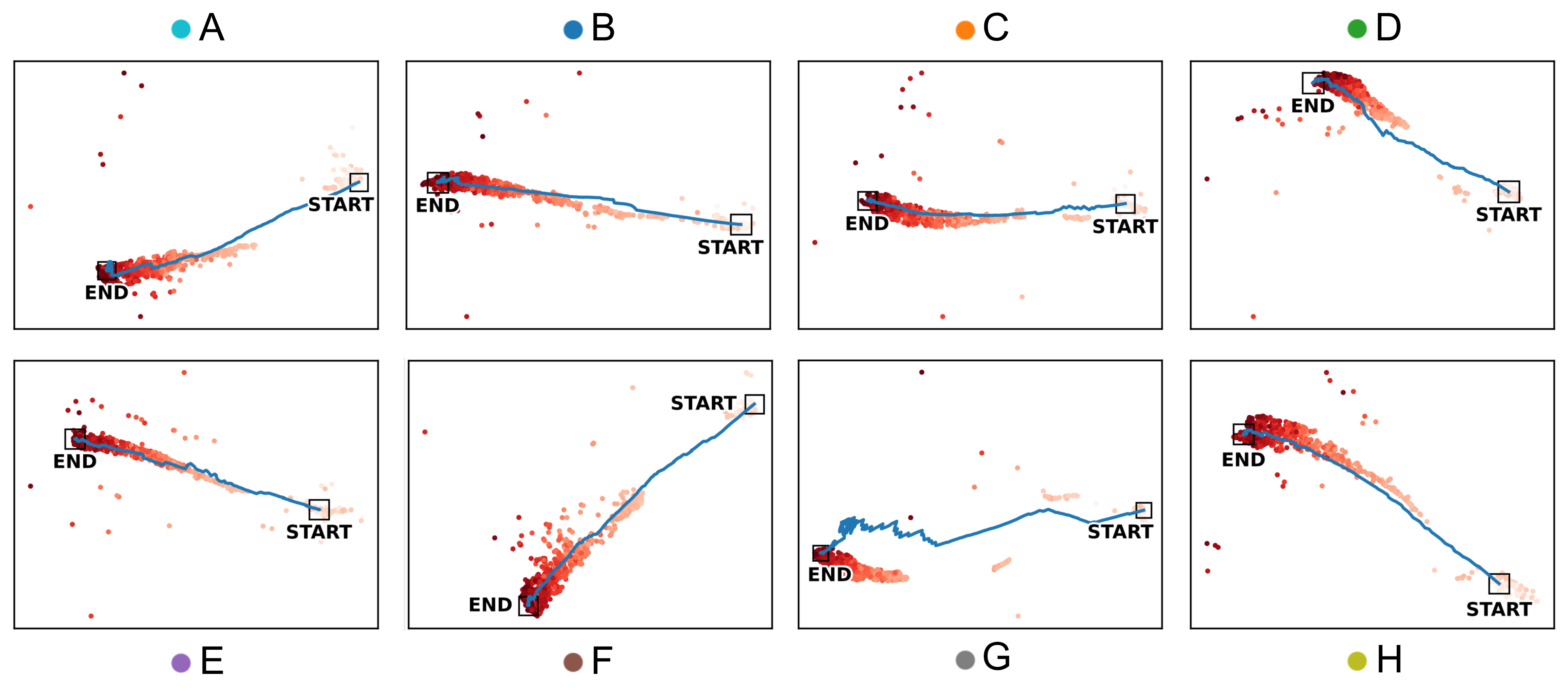}
\caption{Image patch estimated trajectories for each plant variety in the train set, overlaid onto test set. Varieties with larger gaps in latent space are more difficult to model, which introduces noise in some trajectories. However, all trajectories converge to the end of their respective sequence.}
\label{fig:traj_est_per_geno}
\end{figure}

Trajectories modeled based on the training set typically closely follow the data in the validation set, as shown in \cref{fig:traj_est_per_geno}. Exceptions to this can be seen when the projection of a variety has gaps and isn't continuous in latent space. Despite this, estimated trajectories still end up in the correct areas by the end of the time series.

\subsection{Generalization} \label{sec:generalize}
To explore generalization, we consider a TLT module with a class prediction head trained only on a subset of the plant varieties in our dataset, and evaluate performance on unseen plant varieties for patch-based data.

\begin{table}[h!]
\centering
\begin{tabular}{c c c } 
 \hline
  Backbone & Withheld Classes & Class PA* [\%] $\uparrow$ \\ 
 \hline
 DINOv2 & A, B & 91.8\%  \\ 
 DINOv2 & D, G & 75.4\% \\
 SigLIP & A, C, H & 82.8\% \\ 
 SigLIP & F, B, D & 61.3\% \\
 \hline
\end{tabular}
\caption{Image patch class prediction on previously unseen genotypes. SigLIP maintains strong performance, even when three of eight total classes are withheld from training. *Modified PA metric described in \cref{sec:generalize}, which assigns each component in a Gaussian mixture model the class label of the class it primarily contains.}
\label{table:unseen_geno_perf}
\end{table}

\begin{figure}[h!]
\centering
\includegraphics[width=0.82\linewidth]{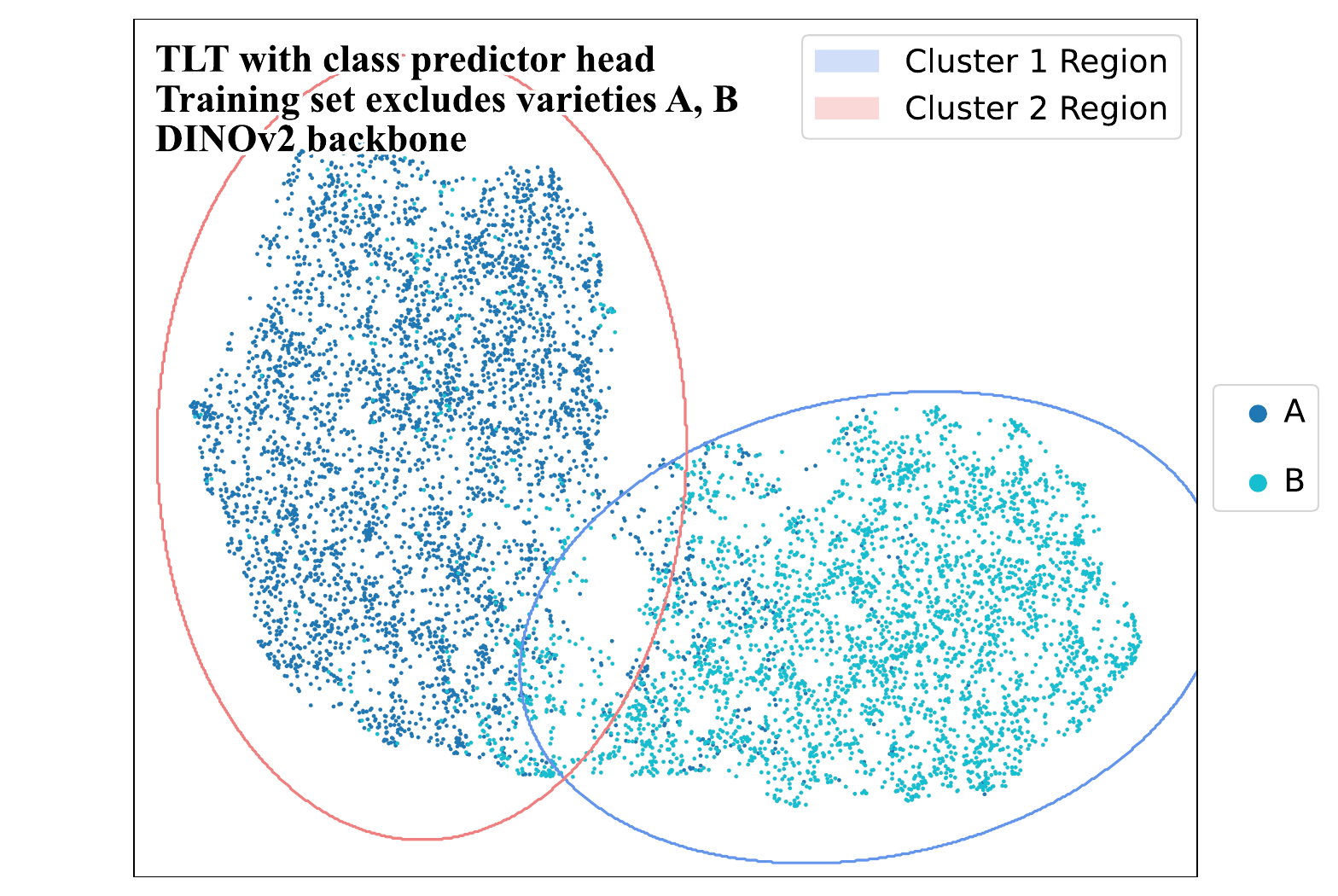}
\caption{Providing the image patch TLT with previously unseen genotypes demonstrates generalization of the encoder, as unseen genotypes still separate in UMAP space. Ellipses depict 95\% confidence region for each component in a 2 component Gaussian mixture model fit to the data.}
\label{fig:umap_unseen}
\end{figure}

When withholding varieties from the training set, we cannot evaluate class prediction in the typical way, as the prediction heads are never trained to predict a variety outside of the training set. Instead, when reserving $N$ varieties for the test set, we fit an $N$ component Gaussian mixture model to the projected test features. We evaluate how well these components delineate varieties by assigning each component the class label corresponding to which class it primarily contains, using ground truth. We then compute percent agreement of classification. Depending on how many classes are withheld, the TLT module is able to strongly separate unseen classes, as seen in \cref{table:unseen_geno_perf} and in Figure \ref{fig:umap_unseen}. TLT modules for the time prediction task generalize strongly to unseen varieties. The performance drop compared to in-class prediction is marginal as seen in \cref{table:unseen_geno_perf_time}.

\begin{table}[h!]
\centering
\begin{tabular}{c c c } 
 \hline
  Backbone & Withheld Classes & Time MAE(d) $\downarrow$ \\ 
 \hline
 DINOv2 & D, G & 3.88 $\pm$ 3.06  \\ 
 DINOv2 & A, G, H & 4.70 $\pm$ 4.29 \\
 SigLIP & A, B & 4.19 $\pm$ 4.17 \\
 SigLIP & A, C, H & 4.704$\pm$4.29 \\ 
 \hline
\end{tabular}
\caption{Image patch time prediction performance on previously unseen genotypes. All backbones have strong performance comparable to in-class predictions.}
\label{table:unseen_geno_perf_time}
\end{table}

\subsection{Explainability} \label{sec:explainability}
Once we have a trained model, we visualize which sections of an image are the most important for performing the pretext tasks. We apply Grad-CAM \cite{selvaraju2016grad} to generate preliminary visualizations of the attention maps in a trained image patch TLT module.

Guided by results in the previous subsections, we train a TLT module with a DINOv2 backbone and a time prediction head. We seek to determine which visual features are most important when predicting time. Using two images from different time points $t_{0}, t_{1}$, we obtain an encoded feature vector for each image. The Grad-CAM method then back-propagates the cosine similarity error between these vectors to the attention maps of the feature extractor backbone. These attention maps are then visualized as heatmaps.

\begin{figure}[h]
    \centering
    \includegraphics[width=0.85\linewidth]{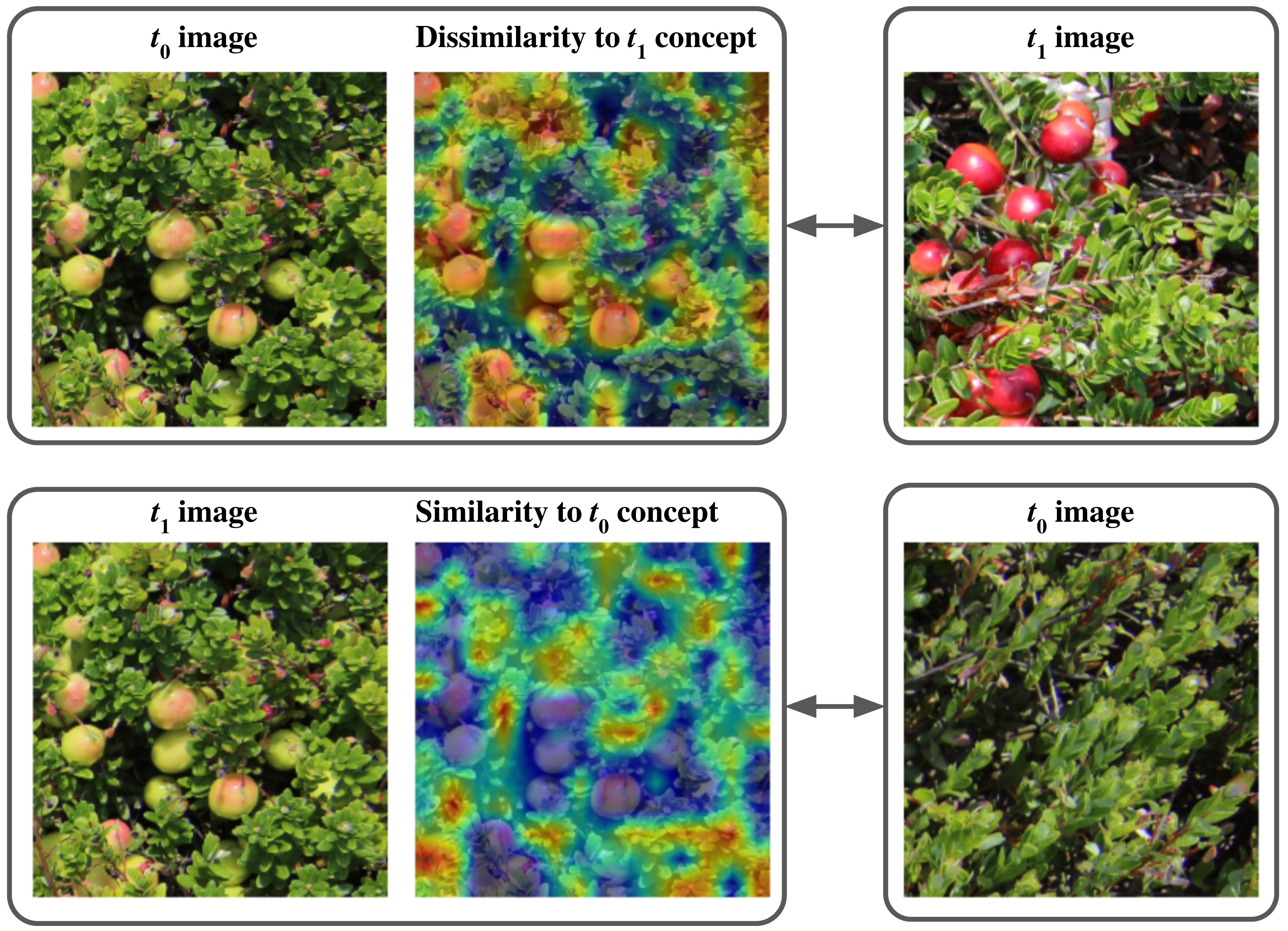}
    \caption{Grad-CAM visualizations suggest that berries serve as a key visual difference to predict time in the image patches.}
    \label{fig:grad-cam-viz}
\end{figure}

In \cref{fig:grad-cam-viz}, we present a visualization that presents the similarity and dissimilarity between two sets of images. In both sets of images, one can observe varying levels of ripeness due to change over time. In the top set of images, Grad-CAM emphasizes the difference in ripeness, highlighting the unripe berries as visual features that distinguish the two images. In the bottom set of images, the similarities are in terms of the branches and leaves of the cranberry plants. Due to the absence of berries in the compared image, the actual berries are not highlighted as similar by Grad-CAM. These results imply that the model is using the ripening and growth of a cranberry plant to estimate time. This suggests potential for using the time prediction pretext task as an self supervised way to detect ripeness. However, it should be noted that these results are still in their initial stages, and that more can be done to verify the robustness of Grad-CAM on our results as a whole.

%% file: sec/6_conclusion.tex
\section{Conclusion}

We present the  time-lapse tracking (TLT) framework that learns latent representations of time-series data from cranberry crop images. The TLT prediction module uses a series of pretext tasks to fine-tune a latent vector that can be used to monitor cranberry growth over a period of time. TLT can be used by breeders and growers to quantitatively nuanced crop development, helping growers maximize their yields with minimal resources and breeders more efficiently screen for superior plant phenotypes. We also provide the Time-lapse Cranberry Dataset (TLC), which contains single view images of eight cranberry varieties over a growing season to support our change monitoring task. 

We evaluate our method quantitatively through our pretext tasks and qualitatively through the visual appearance of the projected features. In terms of quantitative performance, we found that the DINOv2 and SigLIP foundation models tended to be the best performing backbones for our pretext tasks. DINOv2 performed the best in the case of a single time head, but the results were more mixed when we introduced more pretext training heads. In this case, ViT performs the best in the time prediction task, but DINOv2 performs the best in the classification tasks. When we visualize these features, clear separation emerges based on variety and time. Furthermore, we observe that when we withhold a subset of varieties from the training set, the TLT module is able to generalize to these unseen varieties. Overall, our method provides a way for cranberry breeders and growers to comprehensively understand the state of their crops throughout the growing season. 

\label{sec:conclusion}

%% file: sec/7_acknowledgements.tex
\section*{Acknowledgments}
 This project was sponsored by the NSF NRT-FW HTF: Socially Cognizant Robotics for a Technology Enhanced Society (SOCRATES) No. 2021628.